\documentclass{article} 
\usepackage[final]{nips_2016}
\usepackage{times}
\usepackage{hyperref}
\usepackage{amsmath}
\usepackage{amsthm}
\usepackage{graphicx}
\usepackage{caption}
\usepackage{subcaption}
\usepackage{url}
\usepackage{eqparbox}
\usepackage{threeparttable}
\usepackage{multicol}
\usepackage{array}
\usepackage{amssymb}
\usepackage{proof}
\usepackage{algorithm,algorithmic}
\algsetup{linenosize=\small}
\usepackage[font=small,labelfont=bf,margin=0.25cm]{caption}
\usepackage{array}
\usepackage{color}

\newcommand{\cif}{\ensuremath{\mathtt{if}}}

\newcommand{\cobs}{\ensuremath{\mathtt{actual}}}
\newcommand{\cgauD}{\ensuremath{\mathtt{Gaussian}}}
\newcommand{\cchoD}{\ensuremath{\mathtt{Choice}}}
\newcommand{\cbetD}{\ensuremath{\mathtt{Between}}}
\newcommand{\cneaD}{\ensuremath{\mathtt{Near}}}

\newcommand{\Expr}{\ensuremath{\mathit{Expr}}}
\newcommand{\Ref}{\ensuremath{\mathit{Ref}}}

\newcommand{\nop}{\ensuremath{\mathit{op}}}
\newcommand{\nprim}{\ensuremath{\mathit{primOp}}}
\newcommand{\nerp}{\ensuremath{\mathit{erp}}}
\newcommand{\nblack}{\ensuremath{\mathit{blackOp}}}

\newcommand{\dom}{\ensuremath{\mathit{dom}}}
\newcommand{\score}{\ensuremath{\mathit{score}}}

\theoremstyle{definition}

\title{Spreadsheet Probabilistic Programming}

\author{
	Mike Wu  \hspace{.5cm}
	Yura Perov  \hspace{.5cm}
	Frank Wood \hspace{.5cm}
	Hongseok Yang \hspace{.5cm} \\ \\
\{mike,yura,frank,hongseok\}@invrea.com
}

\newtheorem{theorem}{Theorem}[section]

\newtheorem{lemma}[theorem]{Lemma}

\begin{document}

\maketitle

\begin{abstract}
Spreadsheet workbook contents are simple programs.  Because of this, probabilistic programming techniques can be used to perform Bayesian inversion of spreadsheet computations.  What is more, existing execution engines in spreadsheet applications such as Microsoft Excel can be made to do this using only built-in functionality.   We demonstrate this by developing a native Excel implementation of both a particle Markov Chain Monte Carlo variant and black-box variational inference for spreadsheet probabilistic programming.  The resulting engine performs probabilistically coherent inference over spreadsheet computations, notably including spreadsheets that include user-defined black-box functions. Spreadsheet engines that choose to integrate the functionality we describe in this paper will give their users the ability to both easily develop probabilistic models and maintain them over time by including actuals via a simple user-interface mechanism.  For spreadsheet end-users this would mean having access to efficient and probabilistically coherent probabilistic modeling and inference for use in all kinds of decision making under uncertainty.

\end{abstract}

\section{Introduction}
Spreadsheets are the de facto lingua franca of data analysis \citep{panko2008spreadsheet}. They are the principle what-if simulation and decision-making tool for millions of users \citep{scaffidi2005estimating,chan1996use}.  Spreadsheet users often translate internal beliefs and expert domain knowledge into simulations in the form of spreadsheet programs, without necessarily even realising they are programming.  A common spreadsheet simulation is one in which assumptions are set apart, often on a separate worksheet, and a dependent forward-simulation is specified, for example a sequence of dividend payments given a simulation of the finances and decision making policy of a corporate entity.  These simulations are used to make predictions for decision-making under uncertainty;  for instance an investment decision based on the distribution of an internal rate of return calculation.  The usual way stochasticity is injected into such simulations is by manually varying the values of assumptions to reflect uncertainty in held beliefs about their values.  Model checking is implicit as all subcomputations may be plotted and ``eye-balled'' to assess their realism; unrealistic simulators are simply reprogrammed immediately.  Conditioning is manual in the sense that constraining said models to reflect observed actuals relies upon the spreadsheet user manually editing the spreadsheet, replacing previously simulated cells with observed actual values.   Our probabilistic programming approach to spreadsheet modeling introduces a novel approach to this latter procedure via the notion of observation, but remains compatible with existing usage paradigms.

The principle contribution of this paper is the idea that automatic Bayesian model inversion in spreadsheet computation is possible and derives from the connection between spreadsheets, programming languages, and, consequently, probabilistic programming.  The design for introducing the notion of observation in the spreadsheet framework is novel, so too are the algorithms which enable our native implementations.  Our abstract spreadsheet programming language also allows us to formalise connections between language expressivity and inference algorithm formal requirements in a way that  further solidifies the footings of the machine learning probabilistic programming literature \citep{pfeffer2001ibal,GMR+08,pfeffer2009figaro,wingate2011lightweight,Wood-AISTATS-2014,Paige-ICML-2014,vandeMeent-AISTATS-2015},  particularly  that part which advocates  variational inference   \citep{wingate2013automated,mansinghka_arxiv_2014,kucukelbir2014fully}, and particularly black-box variational inference \citep{ranganath2013black}, for probabilistic programming.


\section{Abstract Spreadsheet Language}

We start by formalizing the syntax and semantics of our spreadsheet language
and proving important properties of the language. These properties enable us
to safely employ certain inference algorithms as discussed later in this paper. 

\begin{table}[t]
{\small
$$
\begin{array}{@{}c@{}}
\begin{array}{@{}l@{\ \ }l@{\qquad\qquad}l@{\ \ }l@{\qquad\qquad}l@{\ \ }l@{}}
\mbox{Constant Numbers} & c 
& \mbox{References of Cells} & r
& \mbox{Labels} & l
\\[0.5ex]
\mbox{Primitive Operators} & \nprim
& \mbox{Black-Box Operators} & \nblack
\end{array}
\\[3ex]
\begin{array}{@{}l@{\ \ }l@{\;}c@{\;}l@{}}
\mbox{Expressions} & e & {} ::= {} & 
  c 
  \,\mid\,r 
  \,\mid\,\nop_l(e_1,...,e_n)
  \,\mid\,\cif\, e_1\,e_2\, e_3 
  \,\mid\,\cobs(c,\nerp_l(e_1,...,e_n))
\\[0.5ex]
\mbox{Operators} & \nop & {} ::= {} & 
  \nprim \,\mid\,
  \nblack \,\mid\,
  \nerp 
\\[0.5ex]
\mbox{Elementary Random Proc.}
& \nerp & {} ::= {} &
  \cgauD \,\mid\,
  \cchoD \,\mid\,
  \cbetD \,\mid\,
  \cneaD
\end{array}
\end{array}
$$
}
\caption{Abstract spreadsheet language grammar. \label{table:pl-syntax}}
\end{table}

Intuitively, spreadsheets are finite maps from references of table cells to program
expressions, which specify how to calculate the value of the current cell 
using those of other cells. Table~\ref{table:pl-syntax} shows a grammar for
these expressions $e$ associated with cells. The grammar uses $c$ for constant
numbers (such as $1.0$ and $-2.4$), $r$ for references of cells, 
$\nprim$ for primitive operators (such as $+$ and $\log$), and $\nblack$ for
user-defined black-box operators. Typically these black-box operators are
external custom functions, such as Excel VBA functions, and they may be stochastic and
model unknown probability distributions.  According to the
grammar, an expression can be a constant $c$, the value of a cell $r$,
or the result of applying deterministic or stochastic opertions $\nop_l(e_1,...,e_n)$.
These applications are annotated with unique labels $l$, which we will use to name
random variables associated with spreadsheets. An expression can also be the conditional 
statement $\cif\,e_1\,e_2\,e_3$, which executes $e_2$
or $e_3$ depending on whether the evaluation of $e_1$ gives a non-zero value or not.
The last possibility is the actual statement $\cobs(c,\nerp_l(e_1,\ldots,e_n))$, which
states that a random variable with the distribution $\nerp_l(e_1,\ldots,e_n)$ is observed
and has the value $c$. The $\nerp$ here is also annotated with a unique label $l$.  Note that these labels
will typically not be part of any concrete instantiation of this abstract language, but, due to the 
properties that follow, can easily be added at compile time in a single pass over the spreadsheet.  Also note that $\cobs$ is novel to spreadsheet languages but closely corresponds to the notion of observation in the probabilistic programming literature \cite{mansinghka_arxiv_2014,Wood-AISTATS-2014}.

Let $\Expr$ be the set of all expressions this grammar can generate. Formally, a \emph{spreadsheet} is a finite map
$f$ from references of cells to expressions in $\Expr$. We write
$$
f : \Ref \to \Expr
$$
to denote a spreadsheet $f$ whose domain is $\Ref$. Note that $\Ref$ is finite since $f$
is a finite map. $\Ref$ consists of cells used in the spreadsheet and $f$ describes
expressions associated with these cells.

We say that a spreadsheet $f : \Ref \to \Expr$ is \emph{well-formed} if the following directed graph $G$ with vertex set $V$ and edge set $E$
does not have a cycle:\footnote{Formally, this acyclicity means that the transitive closure $E^+$ of $E$
does not relate any $r \in V$ to itself:
$$
E^1 = E,
\qquad
E^{n+1} = \{(r,r') \,\mid\,(r,r'') \in E\ \mbox{and}\ (r'',r') \in E\ \mbox{for some $r''$}\},
\qquad
E^+ = \bigcup_{n \geq 1} E^n.
$$}
$$ 
G = (V,E),
\qquad\
V = \Ref,
\qquad\
E = \{(r,r') \,\mid\,r,r' \in \Ref \ \mbox{and $r$ occurs in expression $f(r')$}\}.
$$
Intuitively, this acyclicity condition means the absence of a circular dependency among reference cells
in a spreadsheet. In this paper, we consider only well-formed spreadsheets.

\begin{figure}
{\small
$$
\begin{array}{@{}c@{}}
\infer{
  r \Downarrow_\rho \rho(r), (0,0,\emptyset,[])
}{
  r \in \dom(\rho)
}
\qquad
\infer{
  \nprim_l(e_1,....,e_n) 
  \Downarrow_\rho
  c, (w_1 \oplus ... \oplus w_n)
}{
  e_i \Downarrow_\rho c_i, w_i
  \ \mbox{for all $1 \leq i \leq n$}
  &&
  c = \nprim(c_1,...,c_n)
}
\\
\\
\infer{
  \nblack_l(e_1,....,e_n) 
  \Downarrow_\rho
  c, (\sum_i p_i, \sum_i q_i, \bot, \mathit{concat}(L_1,...,L_n,[l])) 
}{
  e_i \Downarrow_\rho c_i, (p_i,q_i,d_i,L_i)
  \ \mbox{for all $1 \leq i \leq n$}
  &&
  c \sim \nblack(c_1,...,c_n)
}
\\
\\
\infer{
  \nerp_l(e_1,....,e_n) 
  \Downarrow_\rho
  c, (w_1 \oplus ... \oplus w_n \oplus (p,q,[l : g],[l]))
}{
\begin{array}{@{}l@{}}
  e_i \Downarrow_\rho c_i, w_i
  \ \mbox{for all $1 \leq i \leq n$}
  \qquad\
  (Q, \lambda) = \mathit{getProposal(l)}
  \qquad\
  c \sim Q(c_1,...,c_n ; \lambda)
\\
  p = \score(\nerp_l(c_1,...,c_n),c)
  \quad\
  q = \score(Q(c_1,...,c_n ; \lambda),c)
  \quad\
  g = \nabla_\lambda \score(Q(c_1,...,c_n;\lambda),c)
\end{array}
}
\\
\\
\infer{
  \cif\ e_1\ e_2\ e_3 \Downarrow_\rho c', w \oplus w'
}{
  e_1 \Downarrow_\rho c, w
  &&
  c \neq 0
  &&
  e_2 \Downarrow_\rho c', w'
}
\qquad
\infer{
  \cif\ e_1\ e_2\ e_3 \Downarrow_\rho c', w \oplus w'
}{
  e_1 \Downarrow_\rho 0, w
  &&
  e_3 \Downarrow_\rho c', w'
}
\\
\\
\infer{
  \cobs(c,\nerp_l(e_1,...,e_n))
  \Downarrow_\rho
  c, w \oplus (w_1 \oplus ... \oplus w_n \oplus (p,0,\emptyset,[l]))
}{
  e_i \Downarrow_\rho c_i, w_i
  \ \mbox{for all $1 \leq i \leq n$}
  &&
  p = \score(c,\nerp_l(c_1,...,c_n))
}
\\
\\
\infer{
  \rho \xrightarrow{p,q,\Lambda,L}_f \rho'
}{
  r\ \mbox{is the $\prec$-least element in $(\Ref \setminus \dom(\rho))$}
  &&
  f(r) \Downarrow_\rho c, (p,q,\Lambda,L)
  &&
  \rho' = \rho[r : c]
}
\end{array}
$$
}
\caption{Rules for deriving evaluation relations for spreadsheets and expressions.
We use $\emptyset$ for the empty finite function, $[]$ for the empty relation,
$\rho[r : c]$ for the update of $\rho$ with new binding of $r$ and $c$,
and $\mathit{concat}(L_1,...,L_n)$ for the concatenation of sequences $L_1,...,L_n$.  Note that $\score$ returns log values. \label{table:pl-semantics}}
\end{figure}

One useful consequence of our well-formedness condition is that we can compute a total order
of all cell references of a spreadsheet that respects the dependency relationship. This can be
achieved by the well-known topological-sort algorithm, which enumerates vertices of a given 
finite directed acyclic graph $(V,E)$ to a sequence $[v_1,v_2,\ldots,v_n]$ such that for every edge $(v,v') \in E$, 
the vertex $v$ appears before $v'$ in the sequence. 
\begin{lemma}
For every spreadsheet $f : \Ref \to \Expr$, there exists an enumeration $[r_1,\ldots,r_n]$ of
all references in $\Ref$ such that for all $r,r' \in \Ref$, if $r$ occurs in $f(r')$, it appears
before $r'$ in the enumeration. This enumeration can be computed by topological sort.
\end{lemma}

Simple yet important properties of well-formed spreadsheets are that they
always terminate and that they use bounded numbers of random variables. These two 
properties enable us to show that such spreadsheets are probabilistic models with acyclic dependencies, and that we can safely perform inference
over spreadsheet calculations using algorithms developed for such models. 
The properties hold because well-formedness bans circular dependency and
expressions used in these spreadsheets do not have loop or recursion.
In the rest of this section, we formally prove these properties. We 
use a fixed well-formed spreadsheet $f : \Ref \to \Expr$,
assume the enumeration $[r_1,\ldots,r_n]$ of $\Ref$ generated by 
the topological sort as described by the previous lemma, and write
$r_i \prec r_j$ for $r_i,r_j \in \Ref$ when $r_i$ appears before $r_j$
in this enumeration.

Define a \emph{state} $\rho$ to be a function from a subset of $\Ref$,
denoted $\dom(\rho)$, to numbers such that 
$$
\forall r,r' \in \Ref.\;
(r \prec r' \wedge r' \in \dom(\rho)) \implies 
r \in \dom(\rho).
$$
A state $\rho$ represents a partially-evaluated spreadsheet, and specifies
the values of evaluated cells. The condition for $\rho$ just means that 
the evaluation occurs according to the total order $\prec$. 

The formal semantics of spreadsheets is defined in terms of two evaluation relations,
one for entire spreadsheets and the other for expressions. Let
$p,q$ real numbers, $\Lambda$ a finite map from labels to real numbers sequencies, $L$ a sequence of labels,
$\rho,\rho'$ spreadsheet states, and $e,e'$ expressions. These
relations have the following forms
$$
\rho \xrightarrow{p,q,\Lambda,L}_f \rho'\qquad\quad\mbox{and}\qquad\quad
e \Downarrow_\rho c,(p,q,\Lambda,L).
$$
The first relates two spreadsheet states $\rho$ and $\rho'$, and describes that evaluating 
$f$ one step from $\rho$ results in $\rho'$. The tuple $(p,q,\Lambda,L)$ is bookkeeping about 
this evaluation: during the evaluation, $|L|$-many values are sampled from applications
with labels in $L$, the total log density of these samples according to their proposal
distributions is $q$, the gradients of the densities of these proposals with respect to their parameters 
form a map $\Lambda$, and the log density of the samples according to
the target joint distributions is $p$. This single step evaluation computes the value of
a cell $r$ so that $\{r\} =  \dom(\rho') \setminus \dom(\rho)$.
The second relation specifies similar information about expressions. It says that $e$ evaluates 
to a number $c$ possibly in multiple steps (rather than in one step), and
that the tuple $(p,q,\Lambda,L)$ records the very information that we have just described for $\rho$, but
this time for this multi-step evaluation of the expression $e$.
The rules for deriving these evaluation relations are given in
Table~\ref{table:pl-semantics}. Each rule says that if the conditions above the
bar hold, so does the statement below the bar. The rule for a reference $r$
says that the expression $r$ gets evaluated by the simple look up 
of the spreadsheet state $\rho$. The bookkeeping part, often denoted by a symbol $w = (p, q, \Lambda, L)$, 
in this case is a tuple of two
zeros, the empty finite function $\emptyset$, and the empty sequence $[]$. According
to its rule, the evaluation of $\nprim_l(e_1,....,e_n)$ first executes all of its parameters
$e_1,...,e_n$ to get $(c_1,w_1),...,(c_n,w_n)$, and then combines these results. $c_i$'s
get combined by the primitive operator $\nprim$, denoted $c$ in the rule, and 
$w_1,\ldots,w_n$ by the $\oplus$ operator that is defined as follows:
$(p,q,\Lambda,L) \oplus (p',q',\Lambda',L') = (p+p', q+q', \Lambda'', L'')$ 
where $L'' = \mathit{concat}(L',L'')$, the concatenation of $L'$ and $L''$,
and 
$$
\Lambda''(\lambda) = 
\left\{\begin{array}{ll}
\Lambda(\lambda) & \mbox{if}\ (\lambda \in \dom(\Lambda)),
\\
\Lambda'(\lambda) & \mbox{else if}\ (\lambda \in \dom(\Lambda')),
\\
\mbox{undefined} & \mbox{otherwise}.
\end{array}\right.
$$
The case of the black-box operator is similar except that the resulting number $c$ 
is sampled according to the operator, the bookkeeping part records the use of
this random variable by adding $l$ to the end of $\mathit{concat}(L_1,...,L_n)$,
and its $\Lambda$ component becomes $\bot$, which represents the absence of information
on gradient. This $\bot$ is an annihilator. When it gets combined with another $\Lambda'$ 
(from both directions) in $\oplus$, the result is always $\bot$. The rule for the
$\nerp$ application is the most complex, but follows the similar pattern. 
According to this rule, the evaluation of $\nerp_l(e_1,...,e_n)$ first runs its arguments
and obtains $(c_1,w_1),...(c_n,w_n)$. Then, it looks up a proposal distribution $Q$
at the label $l$, which has a parameters vector $\lambda$. The evaluation gets a sample $c$
from $Q$, and computes the log densities $p$ of the prior $\nerp(c_1,...,c_n)$
and $q$ of the proposal $Q(c_1,...,c_n ; \lambda)$, as well as the gradient $g$ of 
$Q(c_1,...,c_n;\lambda)$ with respect to $\lambda$. These $p$, $q$, the
singleton map from $l$ to $g$, and the label $l$ are all added to
the bookkeeping of this evaluation.  The meaning of the remaining rules for $\Downarrow_\rho$ follow suit.

We have only one rule for $\rightarrow_f$. It says that 
the evaluation of $f$ at $\rho$ first picks the next unevaluated cell $r$,
then executes the expression stored at $r$, and incorporates the result
$(c,(p,q,\Lambda,L))$ of this execution by associating $r$ with $c$ in $\rho$,
and recording $(p,q,\Lambda,L)$ on top of $\rightarrow_f$.

\begin{table}[t]
{\small
$$
\begin{array}{@{}r@{\;}c@{\;}l@{}}
\multicolumn{3}{c}{
  V =  \Ref
  \qquad
  E = \{(r,r') \,\mid\,\mbox{$r$ occurs in the expression $f(r')$}\}
  \qquad
  G = (V,E)
}
\\[1ex]
  \prec & = & \{(r,r') \,\mid\,\mbox{$r$ appears before $r'$ in the topological sort of $(V,E)$}\}
\\[0.5ex]
  V_O & = & \{ r \,\mid\, \mbox{$f(r) = \cobs_l(c,\nerp(e_1,...,en))$ for some $c,e_i$}\} 
\\[0.5ex]
  [v_{o_1},...,v_{o_n}] & = & \mbox{the sorted list of $V_O$ according to $\prec$}
\\[0.5ex]
  V_{\dashv 1} & = & \{ r \,\mid\, (r_1,r_2),...,(r_m,v_{o_1}) \in E\ 
       \mbox{for a nonempty sequence $[r_1,...,r_m]$}\}
\\[0.5ex]
  V_{\dashv (i+1)} & = & \{ r \,\mid\, (r_1,r_2),...,(r_m,v_{o_{i+1}}) \in E\ 
       \mbox{for a nonempty sequence $[r_1,...,r_m]$}\} \setminus V_{\dashv i}
\\[0.5ex]
  V_{* i} & = & 
  \{ r \,\mid\, (r,r') \in E\ \mbox{for some $r' \in (V_{\dashv i} \cup \{v_{o_i}\})$}\}
  \setminus
  (V_{\dashv i} \cup \{v_{o_i}\})
\\[0.5ex]
  V_r & = & V \setminus \bigcup_{i = 1}^n V_{\dashv i}
\\[0.5ex]
  V_{* r} & = & 
  \{ r \,\mid\, (r,r') \in E\ \mbox{for some $r' \in V_r$}\}
  \setminus
  V_r
\end{array}
$$
}
\caption{Notations for the graph $G = (V,E)$ generated by a well-formed spreadsheet $f:\Ref \to \Expr$.\label{table:pl-notations}}
\end{table}

\begin{theorem}[Termination]
All well-formed spreadsheets terminate. Technically, this means that 
for every well-formed spreadsheet $f$, there is no infinite sequence
$$
\rho_1 \xrightarrow{p_1,q_1,\Lambda_1,L_1}_f
\rho_2 \xrightarrow{p_2,q_2,\Lambda_2,L_2}_f
\rho_3
\qquad\ldots\qquad
\rho_k \xrightarrow{p_k,q_k,\Lambda_k,L_k}_f 
\rho_{k+1}
\qquad\ldots
$$
and that for every state $\rho$ and expression $e$, there is 
no infinite derivation tree with the conclusion $(e \Downarrow_\rho c,w)$
for some $c,w$.
\end{theorem}
This theorem holds because if $\rho \xrightarrow{p,q,\Lambda,L}_f \rho'$,
then $\dom(\rho')$ is strictly larger than $\dom(\rho)$, and
in every rule for $(e \Downarrow_\rho c,w)$, all assumptions are about
subexpressions of $e$ not equal to $e$ itself.

\begin{theorem}[Bounded Number of Random Variables]
Let $f : \Ref \to \Expr$ be a well-formed spreadsheet, and
let $L = \{l \,\mid\, \mbox{$l$ is a label used in $f(r)$ for some $r \in \Ref$}\}$.
Then, there are $|L|$ or less random variables that cover
all random variables used by the executions of $f$.
\end{theorem}
To see why this theorem holds, let
$\rho_1$ be the empty spreadsheet state $\emptyset$. Then,
by the definitions of our evaluation relations,
whenever we have
$$
\rho_1 \xrightarrow{p_1,q_1,\Lambda_1,L_1}_f
\rho_2
\qquad\ldots\qquad
\rho_i \xrightarrow{p_i,q_i,\Lambda_m,L_i}_f 
\rho_{i+1}
\qquad\ldots\qquad
\rho_m \xrightarrow{p_m,q_m,\Lambda_m,L_m}_f 
\rho_{m+1}
$$
for $\dom(\rho_{m+1}) = \Ref$, the concatenation of $L_1,\ldots,L_m$
does not contain any label more than once. Furthermore, 
all of its labels are included in the set $L$ in the theorem. The claim
of the theorem follows from this observation.

Table~\ref{table:pl-notations} establishes notation for an acyclic graph $G = (V,E)$ 
generated by a well-formed spreadsheet $f : \Ref \to \Expr$; $V_O$ for the set of observed vertices, that is, references of cells 
containing actual statements, which are enumerated in sequence $[v_{o_1},...,v_{o_i}]$ 
according to the total order $\prec$ ,and four types of vertex 
sets: $V_{\dashv i}$ and $V_{*i}$ for certain predecessors of the observed vertex $v_{o_i}$, 
$V_r$ for vertices not affecting observed vertices during the evaluation of a spreadsheet,
and $V_{*r}$ for the immediate predecessors of these vertices.  

\begin{algorithm}[ht]
	\caption{Spreadsheet Sequential Monte Carlo\label{algo:smc}}
	\begin{algorithmic}[1]
	{\small
		\renewcommand{\algorithmicrequire}{\textbf{Input:}}
		\renewcommand{\algorithmicensure}{\textbf{Variables:}}
		\REQUIRE program $f : \Ref \rightarrow \Expr$, joint distribution $P$, proposal distribution $Q$, number of particles $S$, graph $G=(V,E)$, subgraphs $V_{O}$, $\left\{ \ V_{\dashv i} \right\}$, $\left\{ V_{\ast i} \right\}$, $V_r$, $V_{\ast r}$.
			\ENSURE state $\rho$, particles weights $\left\{ w_s \right\}_{s=1}^S$, temporary log likelihoods $T$, database of cells values $\left\{D_s : dom(f) \rightarrow im(\rho)\right\}_{s=1}^S$, temporary database $\left\{D^{tmp}_s\right\}$ for resampling.
		\STATE{\it // Step 1 : Compute the first Actual.}
		\FOR{$s = 1$ to $S$}
			\STATE{Reset $\rho = \emptyset$. Set $D_{s} = \emptyset$. $i = 1$. $T_{p }=0; T_{q } = 0 $ }
			\FOR{$r \in V_{\dashv i}$ following the total order}
				\STATE{$\rho \xrightarrow{p,q, \emptyset,L}_f \rho'$ evaluates $r$ s.t. $\{r\} = \dom(\rho') \setminus \dom(\rho)$}
				\STATE{$T_{p} \mathrel{+}= p$; \ $T_{q} \mathrel{+}= q$}; $\rho = \rho'$
			\ENDFOR
			\STATE{$\rho' \xrightarrow{p,q,\emptyset,L}_f \rho''$ evaluates $v_{o_i}$. $T_{p} \mathrel{+}= p$; \ $T_{q} \mathrel{+}= q$; $w_{s} = \exp(T_{p} - T_{q})$}
			\STATE{{\bf for each} $r \in (V_{\dashv i} \cup {v_{o_i}}), \ D_{s}(r) = \rho''(r)$}
		\ENDFOR
		\STATE{\it // Step 2 : Resample and copy particles.}
		\FOR{$s = 1$ to $S$}
			\STATE{$z \sim \textup{categorical}(\operatorname{norm}(\left\{w_s\right\}))$.}
			\STATE{{\bf for each} $r \in \left \{ V_{\dashv i} \cup \left \{ v_{o_{i}} \right \} \right \}, \ D^{\mathit{tmp}}_{s}(r) = D_{s}(r)$ and $w^{tmp}_s = w_s$}
		\ENDFOR
		\STATE{$D = D^{\mathit{tmp}}$. $\left\{w_s\right\} = \left\{ w^{tmp}_s \right\}$. Set all $w_s$ to $\frac{1}{S}\sum_{s=1}^S{w_s}$.}
		\STATE{\it // Step 3 : Compute remaining Actuals.}
		\FOR{$i = 2$ to $\| V_{O} \| - 1$}
			\FOR{$s = 1$ to $S$}
				\STATE{Reset $\rho = \emptyset$. {\bf for each} $r \in V_{\ast i}, \ \rho = \rho[r : D_s(r)]$}
				\STATE{Repeat lines 3--9.}
			\ENDFOR
			\STATE{Repeat Step 2.}
		\ENDFOR
		\STATE{\it // Step 4 : Propagate changes to other latent cells.}
		\FOR{$s = 1$ to $S$}
			\STATE{Reset $\rho = \emptyset$. {\bf for each} $r \in V_{\ast r}, \ \rho = \rho[r : D_s(r)]$}
			\FOR{$r \in V_{r}$ following the total order}
				\STATE{$\rho \xrightarrow{p,q,\emptyset,L}_f \rho'$ evaluates $r$. $\rho = \rho'$}
			\ENDFOR
			\STATE{{\bf for each} $r \in V_{r}, \ D_{s}(r) = \rho'(r)$}
		\ENDFOR
		\STATE{\it // Step 5 : Outpute posterior distribution.}
		\STATE{For some chosen $\bar{r} \in V$, output a histogram given $\left\{ D_{s}(\bar{r}) \right\}_{s=1}^S$.}
		}
	\end{algorithmic}
\end{algorithm}

\section{Spreadsheet Inference}

\begin{algorithm}[ht]
	\caption{Spreadsheet Black-Box Inference\label{algo:vb} (follows Algorithm 1 from \cite{ranganath2013black})}
	\begin{algorithmic}[1]
	{\small
		\renewcommand{\algorithmicrequire}{\textbf{Input:}}
		\renewcommand{\algorithmicensure}{\textbf{Variables:}}
		\REQUIRE program $f : \Ref \rightarrow \Expr$, joint distribution $P$, distribution $Q$, number of particles $S$, graph $G=(V,E)$, convergence constant $\varepsilon$, bound on iterations $t_{\mathit{max}}$, number of samples per iteration $S$, learning rate parameter $\gamma$, number of stochastic operators in the program $L^{erp}_{*}$.
			\ENSURE state $\rho$, free parameters $\lambda(l)$ of the distribution $Q$ for a particular random choice $l$, joint log probability of a particular sample $T_p$, joint log probability $T_q$ for variational distributions $Q$, vector of gradients for $Q_l$, number of applications for a random choice $T_t(l)$, learning rate $\eta$, change $\Delta\lambda(l)$ in $\lambda(l)$ for a random choice $l$, matrices $G(l)$ for AdaGrad algorithm.
		
		\STATE {\bf for each} label $l \in L^{erp}_{*}$ {\bf do}
		\STATE \ \ \ \ Change $erp_{l}$ to the respective distribution $Q_l$ with $n$ parameters $\lambda(l)$.
		\STATE { \ \ \ \ Initialize $\lambda(l) = {\boldsymbol 0}$. $G(l) = {\boldsymbol 0}.$}
		\STATE {\bf end for each}
		
		\STATE{Set $t = 0$.}
		\REPEAT{}
			\STATE{$t = t + 1$}
			\FOR{$s = 1$ to $S$}
				\STATE{$T_p = 0; T_q = 0$. {\bf for each} $l \in L^{erp}_{*}$, $T_{\Lambda}(l) = {\boldsymbol 0}$ and $T_{t}(l) = 0$.}
				\STATE{Reset $\rho = \emptyset$.}
				\FOR{$r \in V$ in the sorted total order}
					\STATE{$\rho \xrightarrow{p,q,\Lambda,L}_f \rho'$ evaluates $r$ s.t. $\{r\} = \dom(\rho') \setminus \dom(\rho)$}
					\STATE {\bf for each} label $l \in L$ {\bf do}
						\STATE { \ \ \ \ $T_{\Lambda}(l) \mathrel{+}= \Lambda$; $T_t(l) \mathrel{+}= 1$ }
					\STATE {\bf end for each}
					\STATE{$T_{p} \mathrel{+}= p$; $T_{q} \mathrel{+}= q$; $\rho = \rho'$}
				\ENDFOR
			\ENDFOR
			\STATE{$\lambda^{prev} = \lambda$}
			\STATE{{\bf for each} $l \in L^{erp}_{*}$ s.t. $T_t(l) > 0$ {\bf do} }
			\STATE{ \ \ \ \ $\Delta\lambda(l) = \frac{1}{T_{t}(l)} T_{\Lambda}(l) \cdot (T_{p} - T_{q})$ ; $G(l) \mathrel{+}= \Delta\lambda(l) \otimes \Delta\lambda(l)$ }
			\STATE{ \ \ \ \ $\eta = \gamma \ \textup{diag}(\sum_{i=1}^{t} G(l))^{-\frac{1}{2}}$; $\lambda(l) = \lambda(l) + \eta \Delta\lambda(l) $}
			\STATE{{\bf end for each}}
		\UNTIL{$\| \lambda - \lambda^{prev} \|_{2} < \varepsilon$ or $t > t_{\mathit{max}}$ }
		\STATE{For some chosen $l \in L_*^{erp}$, return $q(\lambda(l))$.}
		}
	\end{algorithmic}
\end{algorithm}

Having proven that a spreadsheet terminates and knowing that there exists a total order for the cells in a spreadsheet, we can safely employ algorithms based on sequential Monte Carlo (SMC) for posterior inference over execution paths of spreadsheet programs written in our spreadsheet language.

Algorithm~\ref{algo:smc} gives a detailed implementation of a version of SMC, the inner loop of the particle independent Metropolis Hastings (PIMH)-like algorithm \cite{andrieu2010particle} we implemented in the Excel spreadsheet engine.  
Our SMC algorithm relies the Excel engine to provide $\rho \xrightarrow{p,q,\emptyset,L}_f \rho'$, namely,  
to compute the value $c$ of a 
new cell $r \in \dom(\rho') \setminus \dom(\rho)$ and log scores $p, q$ of
$\nerp$ and proposal distributions respectively.  The trick is to make it do so repeatedly for all particles for observations cells $v_{o_j}$ and their corresponding preceding cells $V_{\dashv j}$ preserving the total order. We resample particles after each evaluated observation. By nature of resampling, particles are not independent and semantically need to be evaluated ``in parallel.''  Our implementation is single-threaded and simulates parallelisation by switching between different states of $\rho$. For every 
spreadsheet state $\rho$ obtained in this repeated evaluation and selected references $r$, the algorithm
stores and reuses $\rho(r)$, if $r$ is in $\dom(\rho)$, in a database $D^{s}(r)$ that is indexed by 
particle number and cell reference.

Up to the first cell containing an observation expression references are evaluated according to the total order, likelihoods are incorporated into weights, and bindings are saved into the database (Alg.~\ref{algo:smc} lines 3--9).  After the first observation, the weights are normalized, and the stored bindings $D$ are resampled accordingly.  For the rest of the observations $v_{o_2}, \ldots, v_{o_{|V_O|}}$, the same procedure is repeated with the exception that directly preceding cells $V_{* j}$ for the cells we need to evaluate $V_{\dashv j}$ must be restored to the state to ensure the evaluation of $v_{o_j}$. This is done by rebinding the references based on values stored in $D$ (Alg.~\ref{algo:smc} lines 19--23). After the observations, changes in each particle are propagated to $V_{r}$ (Step 4). Lastly, the posterior distribution for a reference $\bar{r}$ can be estimated with the final values stored in $\left\{D^s\right\}_{s=1}^S$.

We do not run  sequential Monte Carlo once, but instead we do $M$ independent SMC runs with $S_1, \ldots, S_M$ particles. This improves particles diversity and helps with the problem of sample impoverishment. In order to join these independent SMC islands   $\hat{P}_1 = \sum_{s=1}^{S}{w^1_s \delta_{D^1_s}{(D)}}, \hat{P}_2 = \sum_{s=1}^{S'}{w^2_s \delta_{D^2_s}{(D)}}, \ldots$ into  an unbiased posterior approximation, we weight our isolated particle filters by their evidence estimates $\hat{Z}_j=\frac{1}{S^j}\sum_{s=1}^{S^j}{w^j_s}$ which are saved for each SMC run. 
 That we can do this follows directly from the PIMH results in \citep{andrieu2010particle} where instead of doing MH on ratios of evidence estimates we simply do importance sampling with weights proportional to the evidence estimates.

Additionally, knowing that the spreadsheet graphical model has a finite number of random variables allows us to implement black-box variational inference (BBVI). For each random primitive $l$ in the spreadsheet, BBVI associates a mean field approximation factor $Q_l$. In our implementation we provide normal $\operatorname{Gaussian}(\mu, \sigma)$, uniform continuous $\operatorname{Between}(a, b)$, categorical $\operatorname{Choice}((val_1, \ldots, val_n), (p_1, \ldots, p_n))$ and one-parameter distribution $\operatorname{Near}(val > 0) \sim \operatorname{Gaussian}(val, 0.1 * val)$ random variables. To approximate them, we use variational families $\operatorname{Gaussian}(\lambda_1, \exp{\lambda_2})$, $(b - a) \operatorname{Beta}(\exp{\lambda_1}, \exp{\lambda_2}) + a$, $\operatorname{Choice}((val_1, \ldots, val_n), \frac{1}{\sum_{i=1}^{n}{\exp{\lambda_i}}}(\exp\lambda_1,\ldots,\exp\lambda_n))$ and $\operatorname{Gaussian}(\lambda_1, \exp{\lambda_2})$ correspondigly. Algorithm~\ref{algo:vb} describes black-box inference in further detail.

\section{Experiments}
\label{Experiments}
To demonstrate practicality of implementing both SMC and BBVI inference natively in a spreadsheet  engine, we demonstrate correctness via a regression example and show that both can perform inference over spreadsheets that include user-defined functions. The Excel-syntax abstract spreadsheet language implementation allows users to use random primitives (=GAUSSIAN$(\cdot)$, =CHOICE$(\cdot)$, and =BETWEEN$(\cdot)$) and ``observe'' cells via a syntax =ACTUAL(data, model, parameters) shown in each of the examples that follow.  SMC and BBVI are both implemented in VB and are deployed as Excel Add-In's, meaning that inference functionality can be added to existing spreadsheets. For the examples below, all SMC runs used 5000 particles in islands of 500 and all BBVI runs used 10 samples with 1000 iterations.

\begin{figure}[tbp]
    \tiny
    \begin{minipage}{.4\linewidth}
      \centering
        \includegraphics[scale=0.65]{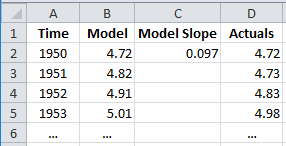}
    \caption{Infered values for GDP example        \label{regression1}
}
    \end{minipage}%
    \begin{minipage}{.6\linewidth}
      \centering
      \includegraphics[scale=0.65]{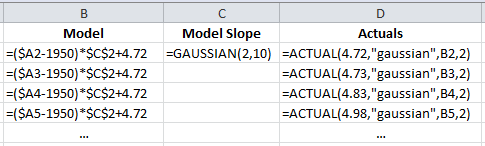}
    \caption{Formulas for GDP example. Columns A and B are ommitted since those cells do not contain formulas.}
    \label{regression2}
    \end{minipage} 
\end{figure}

To illustrate correctness and our novel Excel syntax Figure~\ref{regression2} shows an Excel regression model for US GDP growth versus years from 1950 to 1983.   Figures~\ref{regression1} and \ref{regression2} show the highest probable values of a selection of cells after inference, and their formulas respectively. Notice that overlapping cells in columns B, C and D imply that the formulas in Figure~\ref{regression2}  underly the cells with the same labels in Figure~\ref{regression1}.  This manner of displaying Excel formulae and values is used throughout the experiments section.

The estimated posterior distributions for the slope of the linear model for SMC and BBVI in comparison to the ground truth (GT). BBVI and the ground truth distributions are close to identical ($\mu_{BBVI} = 0.098$, $\mu_{GT} = 0.099$, $\sigma_{BBVI} = 0.019$, $\sigma_{GT} = 0.018$) and SMC offers a good approximation (yielding $0.098$ for the posterior mean slope). 

\begin{figure}[tbp]
  \tiny
    \begin{minipage}{.4\linewidth}
      \centering
        \includegraphics[scale=0.65]{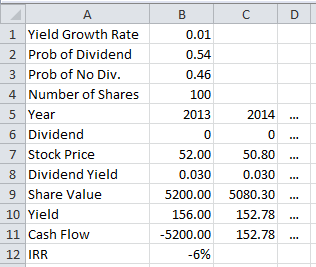}
    \end{minipage} 
    \begin{minipage}{.6\linewidth}
      \centering
        \includegraphics[scale=0.65]{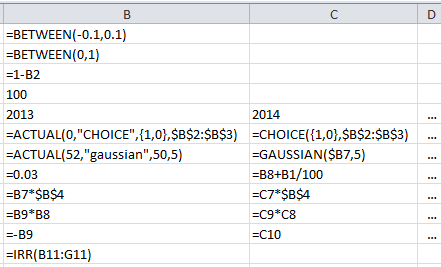}
    \end{minipage} 
    \caption{Cell values (left) and underlying formulas (right) for the IRR example model.}
    \label{figure:irr}
\end{figure}

Another example, shown in Figure~\ref{figure:irr}, illustrates the use of a $blackOp$ primitive, here actually the IRR (internal rate of return) function in Excel, to perform the kind of analysis suggested in the introduction, namely to make an invest or not decision based on the IRR of a cash flow arising from dividend yields, and stock price movements. Because IRR hides an optimization it serves as the kind of $blackOp$ primitive for which no knowledge is available to the spreadsheet about its internal workings. From an end-user perspective, supporting inference over programs with such primitives is important since a significant portion of Excel functionality comes from custom user-defined functions and other similar $blackOp$ functions.  Although not shown, one inference objective that we could compute is to examine the distribution of the IRR in B12 given the effect of observing a dividend in B6 and a stock price in B7 under the specified model.

While our experiments were performed using a prototype, which is quite slow, the approach can be integrated inside any spreadsheet engine since, as shown, it uses only built-in functionality of such engines. By doing such software engineering effort, inference will be much faster and users will be able to run models with many actuals.

\section{Related Work}
\label{Related Work}

Existing work on using variational Bayes \cite{wingate2013automated,mansinghka_arxiv_2014} and specifically black box variational Bayes in probabilistic programming \cite{kucukelbir2014fully} inspired this work.  The formalism we introduce in this paper provides some theoretical justification for some of this prior art but also raises questions, particularly having to do with stochastic optimization in infinite dimensions. 

Here both BBVI and SMC rely upon repeated re-execution of the program guided by proposal distributions.  Having proved termination of all programs written in our abstract spreadsheet language means that we can be assured that the computation performed in the inner loop of our inference algorithm will terminate every time, and, as a result, we can rely on our inference algorithm to terminate too.  Prior probabilistic programming inference work reposed on SMC, notably \citep{Wood-AISTATS-2014,Paige-ICML-2014,mansinghka_arxiv_2014,vandeMeent-AISTATS-2015}. 

There are discernible differences between our approach and Tabular, a probabilistic programming language for Excel created by Microsoft Research \citep{gordon2014tabular}.  Tabular is similarly restricted to our abstract spreadsheet language in the sense that the random choices made in all possible execution paths are finitely enumerable.  The most significant difference between our approach and that of Tabular is that the latter sits ``on the side'' of Excel with execution of its supported inference algorithms performed by a separate runtime, not the Excel engine itself.   Furthermore, Tabular does not allow black-box user-programmed primitives owing to their incompatibility with, for instance, EP \citep{minka2001family} inference (aka Infer.Net \cite{minkainfer}).  We note, however, that progress towards support for black-box factors in EP is in motion \citep{heess2013learning,jitkrittumjust}.

\section{Discussion}
\label{discussion}
We have demonstrated that Bayesian model inversion via both sequential Monte Carlo and black box variational inference are natively implementable in a spreadsheet engine and, moreover, safe in the sense of being on theoretically sound footing.  Implementation in additional spreadsheet engines is ongoing work.   This could bring about a transition from deterministic to probabilistic, conditioned spreadsheet computation which, in turn, could fundamentally impact the way spreadsheets are developed and used for data analysis and modeling in the future.

\bibliographystyle{plainnat}
\small{
\bibliography{refs}
}

\end{document}